\documentclass{article}

\usepackage[dblblindworkshop, final]{neurips_2026}

\workshoptitle{TAE (Trust-AI-Eval): Can We Trust AI Evaluation?}

\usepackage[utf8]{inputenc}
\usepackage[T1]{fontenc}
\usepackage{hyperref}
\usepackage{url}
\usepackage{booktabs}
\usepackage{amsfonts}
\usepackage{amsmath}
\usepackage{microtype}
\usepackage{graphicx}

\title{Auditing VLM Benchmarks from Existing Evaluation Runs}

\author{%
  Denis Degtiarev \\
  Innopolis University \\
  \texttt{d.degtiarev@innopolis.university}
  \And
  Anna Gromova \\
  Innopolis University \\
  \texttt{an.gromova@innopolis.university}
  \And
  Sergey Katkov \\
  Innopolis University \\
  \texttt{s.katkov@innopolis.university}
  \And
  Artem Mikhailin \\
  Innopolis University \\
  \texttt{a.mikhailin@innopolis.university}
  \And
  Safina Mukhamedshina \\
  Innopolis University \\
  \texttt{s.mukhamedshina@innopolis.university}
  \And
  Albert Nasybullin \\
  Innopolis University \\
  \texttt{a.nasybullin@innopolis.university}
  \And
  Karina Shaikhutdinova \\
  Innopolis University \\
  \texttt{k.shaikhutdinova@innopolis.university}
  \And
  Artem Tuzov \\
  Innopolis University \\
  \texttt{a.tuzov@innopolis.university}
}

\begin{document}

\maketitle

\begin{abstract}
Every evaluation run of a vision-language model (VLM) already produces
per-question artifacts (predictions, judged correctness, metadata) that
leaderboards reduce to a single score and then discard. We reuse those
artifacts to audit the benchmark itself, and add an audit layer to
VLMEvalKit: six detectors that run as post-processing over saved outputs,
plus one additional text-only rerun with visual inputs removed, emitting a
per-question report in a single command. The layer reports per-question
findings together with aggregate audit statistics; for the headline
quantities, it attaches permutation-null excess where applicable, interval
estimates for proportions, and sensitivity to the model pool. Auditing
MMStar, a benchmark curated specifically so that the image is
indispensable, with three models, we find zero aggregate accuracy gain
from the image on 20.2\% of its questions. An item-matched permutation
null accounts for 17.0\%, leaving a $+3.2$\,pp excess (corrected
permutation $p=0.001$). This excess is concentrated in the mathematics and
science categories and is absent from coarse perception, a localization
consistent with knowledge substituting for visual evidence; confirming
that mechanism would require item-level review. A second detector flags
134 questions on which every model that answered in parseable form picked
the same non-gold option, and these candidates peak in the perception
categories, whereas the zero-gain excess is largest in mathematics and
science \& technology. The contrasting category profiles show that the two
detectors capture distinct signals. Replacing the LLM judge with exact
string matching changes no conclusion, and every scoring discrepancy
between the two traces to one model whose free-form answers a parser
cannot read. On a binary yes/no benchmark the detectors measurably fall
out of calibration, a scope boundary that we measure and report. The audit
layer, the prediction artifacts and the generated reports accompany this
submission.
\end{abstract}

\section{Introduction}
\label{sec:intro}

Multimodal benchmarks are the instruments through which progress in
vision-language models is measured, compared and rewarded. The scores they
produce steer model development and public leaderboards, yet the
benchmarks themselves are seldom measured with comparable care. A
multiple-choice suite can contain questions answerable without looking at
the image, gold labels that are wrong, and option positions that reward a
guessing strategy, and a model's reported accuracy absorbs all of these
defects silently. Checking for them is usually treated as a separate
research effort, either by building a cleaner replacement benchmark or by
running a dedicated flaw-mining pipeline, each with its own annotation
budget and inference bill, so in practice such checks happen rarely and
cover few benchmarks. Yet every evaluation run already produces the raw
material for exactly this kind of check: per-question predictions, judged
correctness and metadata that leaderboards show as a single score. This paper asks how far that discarded material can be
pushed as an audit of the benchmark itself.

Our case study is MMStar, a benchmark built by filtering candidate
questions so that the image is essential: items solvable by blind
language models were removed by construction. We run three VLMs on it
twice, first as intended and then with every image removed from the
prompt, and align the two runs question by question. On 303 of 1{,}500
questions (20.2\%) the pool is exactly as accurate without the image as
with it. Most of that share is plain arithmetic: even blind answers
unrelated to the items sometimes tie full-run accuracy, and an
item-matched permutation null puts that baseline at 17.0\%. The remaining
$+3.2$\,pp ($p=0.001$) is a paired zero-gain signal, and it concentrates
in the two categories where non-visual knowledge is a plausible
explanation. Finding this required no new benchmark and no annotation
campaign. The audit reuses the full-run artifacts and adds one text-only
inference pass per model; this pass avoids visual processing but is not
cost-free.

We implement this idea as an audit layer inside a mainstream evaluation
toolkit. Six detectors run as post-processing over the per-question files
the toolkit already writes; the only added inference is the text-only
rerun. The layer pairs each headline quantity with the statistics needed
to read it conservatively, and the case study is designed to probe the
audit's own blind spots as well as the benchmark's.

\subsection{Contributions}
\label{sec:contributions}

\begin{enumerate}
  \item \textbf{The first benchmark-audit layer inside a mainstream VLM
  evaluation toolkit.} We extend VLMEvalKit with six benchmark-pathology
  detectors and a toolkit-wide text-only blind-run mode; a single command
  turns saved artifacts from a completed evaluation into a
  per-question audit report with uncertainty statistics and a
  reproducibility manifest. To our knowledge, no mainstream VLM evaluation
  toolkit previously included benchmark-quality checks
  (Section~\ref{sec:method}).
  \item \textbf{A statistical protocol that separates audit signal from
  accuracy arithmetic.} We report every headline quantity as an excess
  over an item-matched permutation null, with interval estimates,
  leave-one-model-out ranges and category strata. On MMStar this protocol
  turns a raw 20.2\% zero-gain share into a defensible $+3.2$\,pp excess
  ($p=0.001$) and exposes how unstable the raw share alone is: it varies
  by 14.2\,pp across model pools while the excess varies by 4.6\,pp
  (Section~\ref{sec:method-stats}).
  \item \textbf{The first published item-level audit of MMStar, with
  quantified validity limits.} We localize two distinct signals inside a
  benchmark curated for vision-indispensability: the zero-gain excess
  concentrates in mathematics and science \& technology, while 134
  unanimous label-conflict candidates concentrate in the perception
  categories, at a rate consistent with an independent re-annotation
  study. The audit also measures its own limits: every conclusion survives
  removal of the LLM judge, and binary yes/no formats fall measurably
  outside calibration
  (Sections~\ref{sec:res-textonly}--\ref{sec:res-binary}).
\end{enumerate}

The pathologies these detectors target are documented at scale by prior
and concurrent work (Section~\ref{sec:related}); our claims concern the
infrastructure, the statistical protocol and the item-level findings. One
measurement
appears to be new outright: Section~\ref{sec:res-judge} quantifies how the
answer-extraction protocol shifts audit conclusions themselves, where
prior work measured its effect on model accuracy.

\section{Related work}
\label{sec:related}

\paragraph{Benchmark pathologies.} Language priors in visual question
answering were established by \citet{goyal2017vqa}; MMStar
\citep{mmstar2024} and MMEvalPro \citep{mmevalpro2024} showed that blind
language models score far above chance on multimodal multiple-choice suites;
concurrent work probes the visual dependence of existing suites through blur
ablations \citep{visualignorance2026} or phrasing-controlled items
\citep{seeorguess2026}; HallusionBench \citep{hallusionbench2024} builds
yes/no items around visual illusions and language priors to elicit these
failure modes deliberately. Label errors destabilize rankings
\citep{northcutt2021pervasive}; re-annotation efforts include MMLU-Redux
\citep{mmluredux2024} and platinum benchmarks \citep{vendrow2025platinum},
and \citet{infodensity2025} report a fallibility analysis covering MMStar.
Position and selection bias in multiple choice is established for language
models \citep{zheng2024selectors}, handled in MMBench through CircularEval
\citep{mmbench2024}, and reused as a contamination probe
\citep{mmdetect2024}. \citet{kim2025correlated} show that model errors
correlate across vendors and scales, which caps what any cross-model
consensus can establish; we return to this in
Sections~\ref{sec:res-consensus} and~\ref{sec:limitations}.

\paragraph{Audit pipelines and toolkits.} The standard responses to these
pathologies treat auditing as a separate research project with its own
inference budget, either by building cleaner replacement benchmarks
\citep{mmstar2024,naturalbench2024,vendrow2025platinum,datbench2026} or by
running standalone flaw-mining pipelines. Fantastic Bugs
\citep{fantasticbugs2025} mines response patterns to flag invalid questions
in nine text benchmarks; BenchMarker \citep{benchmarker2026} applies an
education-inspired rubric with LLM judges to text multiple-choice
benchmarks; DatBench \citep{datbench2026} audits 33 VLM datasets and
releases a cleaned suite. Our deliverable differs: audit machinery embedded
in a mainstream toolkit, reusable across benchmark-model combinations using
artifacts the toolkit already saves. To our knowledge no such layer exists
for VLMEvalKit \citep{vlmevalkit2024} or LMMs-Eval \citep{lmmseval2024}.
Blind-solvability and label errors in VLM benchmarks are documented at
scale by concurrent work \citep{datbench2026,visualignorance2026}; the
checks themselves are therefore established, and this paper moves them
inside the evaluation infrastructure. Redundancy analyses over existing
score matrices \citep{redundancy2025} share the reuse-the-exhaust spirit
for a different question. On the measurement side,
\citet{rosenthal2025unexplored} document protocol sensitivity of VQA
accuracy; we extend that question from accuracy numbers to audit
conclusions.

\section{Methodology}
\label{sec:method}

\begin{figure}[t]
  \centering
  \includegraphics[width=\linewidth]{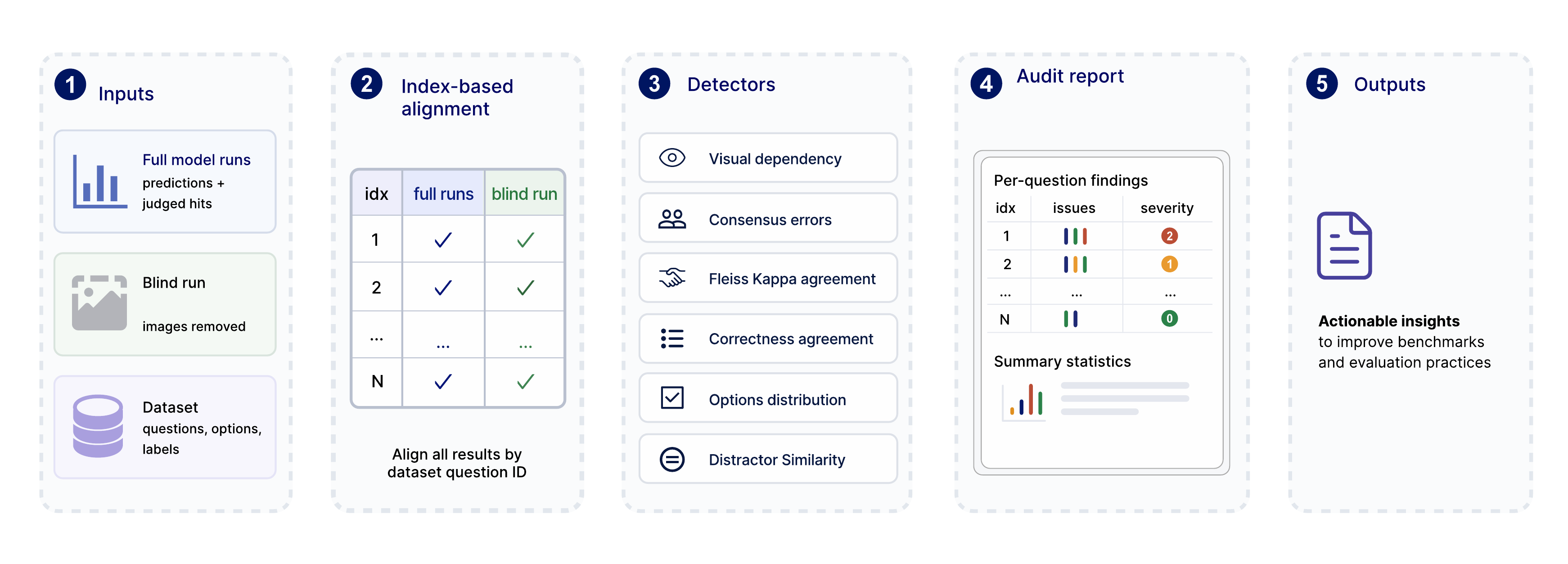}
  \caption{The audit pipeline in five stages. \textbf{(1)}~Inputs: the
  full model runs with predictions and judged hits, one additional
  text-only blind run with images removed, and the dataset itself with
  questions, options and gold labels. \textbf{(2)}~All frames are aligned
  by dataset question ID. \textbf{(3)}~Six detectors process the aligned
  frame. \textbf{(4)}~The audit report lists per-question findings with
  severity levels and summary statistics, including permutation-null
  excess values and a reproducibility manifest. \textbf{(5)}~The report yields
  actionable insights for improving benchmarks and evaluation practice.
  The blind run is the only step that requires inference, and it runs
  without images.}
  \label{fig:pipeline}
\end{figure}

\subsection{Inputs and the blind-run protocol}
\label{sec:method-inputs}

Figure~\ref{fig:pipeline} gives an overview of the pipeline.
The layer consumes the per-question result files of $n$ models on one
benchmark, each with a raw prediction, the gold answer, and optionally a
judged correctness flag. A \emph{blind run} repeats inference with visual
content removed and textual content unchanged. The standard local and API
wrapper paths share the same stripping function; blind artifacts get a
\texttt{\_blind} suffix and sit beside the full run. The toolkit audit
pipeline aligns frames on the dataset-level question ID and restricts them
to the shared ID set.

Two rules protect validity. First, model answers are never back-filled from
gold labels, whatever the judge decided; an earlier version of our own
pipeline made exactly this mistake, and repairing it multiplied the
unanimous-conflict count by eleven (Section~\ref{sec:res-selfaudit}).
Second, predictions from which no option letter can be parsed become an
abstention sentinel that is excluded from consensus votes instead of being
silently counted as wrong.

\subsection{Detectors}
\label{sec:method-detectors}

Table~\ref{tab:detectors} summarizes the six detectors. Consensus,
agreement, position-distribution and distractor detectors parse option
letters from raw predictions and never consult a judge; visual dependency
and correctness agreement consume judged labels and are re-checked under
exact matching in Section~\ref{sec:res-judge}.

\begin{table}[t]
  \caption{Detector taxonomy. ``Judge-dep.'' marks detectors whose output
  depends on the answer-extraction protocol.}
  \label{tab:detectors}
  \centering
  \small
  \setlength{\tabcolsep}{4.5pt}
  \begin{tabular}{lllc}
    \toprule
    Detector & Signal & Pathology flagged & Judge-dep. \\
    \midrule
    \texttt{visual\_dependency} & full vs.\ blind accuracy & no-gain-from-image items & yes \\
    \texttt{consensus\_error} & majority vote vs.\ label & label-conflict candidates & no \\
    \texttt{fleiss\_kappa\_agreement} & inter-model $\kappa$ & unstable / ambiguous items & no \\
    \texttt{correctness\_agreement} & judged-correctness profile & difficulty / discrimination & yes \\
    \texttt{answer\_options\_distribution} & label-position histogram & position bias & no \\
    \texttt{distractor\_similarity} & pairwise option similarity & near-duplicate options & no \\
    \bottomrule
  \end{tabular}
\end{table}

For visual dependency, each question is categorized from model-averaged full
accuracy $a^F_q$ and blind accuracy $a^B_q$: \emph{visual-dependent}
($a^F_q \ge 0.99$ and $a^B_q \le 0.01$), \emph{conflicting}
($a^B_q > a^F_q + \delta$, the image hurts), \emph{visual supplement}
($a^F_q - a^B_q \ge \delta$), and \emph{text-only} otherwise, with
$\delta = 0.2$. With $n=3$ models the accuracies live on the grid
$\{0, 1/3, 2/3, 1\}$, so text-only reduces to exact equality
$a^F_q = a^B_q$ and carries no tunable threshold. The visual-dependent
share, in contrast, is threshold-fragile: it is 22.5\% under the strict
definition and 54.8\% under a majority definition
(Appendix~\ref{app:sweep}), which is why we never headline it. Consensus
conflicts are tiered: \emph{unanimous} (every parseable answer is the same
non-gold option), \emph{high} (all models disagree with the gold label),
\emph{medium} (a qualified majority disagrees).

\subsection{Audit statistics based on the excess over a permutation null}
\label{sec:method-stats}

A raw zero-gain share is easy to over-read. Even if blind answers were drawn
independently of the items, some questions would tie full-run accuracy by
chance, and that baseline moves with the pool's marginal accuracy; across
our two- and three-model pools the raw share varies by 14.2\,pp. The layer
therefore reports the share's \emph{excess over a permutation null}: blind
score vectors are shuffled across questions, with the same permutation
applied to every model, which preserves each model's marginal blind accuracy
and the inter-model correlation structure while destroying only the
per-question full-to-blind pairing. We use 999 permutations; the one-sided
$p$-value uses the standard $(g{+}1)/(N{+}1)$ correction, so the smallest
reportable value is $p=0.001$. The same module attaches Wilson 95\%
intervals to headline proportions (item-level uncertainty),
leave-one-model-out pools (pool sensitivity), a split-half reproducibility
check, and a stratification by official dataset categories (localization and
internal replication). All of it is post-processing without any model use.

\subsection{Evaluation protocol}
\label{sec:method-protocol}

\textbf{Benchmarks.} MMStar: 1{,}500 four-option
multiple-choice items curated for vision-indispensability, with six official
categories of 250 items each. HallusionBench: 951
yes/no items built around visual illusions and language priors, used here as
a deliberate out-of-format probe of the detectors themselves.

\textbf{Models.} InternVL3-8B, Ristretto-3B and gpt-5-nano ($n=3$): two open
models and one API model. This case study is sized to validate the audit
machinery; leave-one-model-out ranges are reported throughout so that
pool sensitivity stays explicit (Table~\ref{tab:pools}).

\textbf{Scoring.} The judged arm uses \texttt{gpt-4o-mini} as the
answer-extraction judge, identical for full and blind runs; the exact arm
parses option letters directly from predictions with no model in the loop.
Consensus, $\kappa$, position and distractor detectors use the exact arm by
construction. Both arms are reported wherever they can differ.

\textbf{Reproducibility.} Every run records its git commit, command-line
arguments and generation parameters in a manifest; the audit is
deterministic given the artifacts and a fixed permutation seed. The audit
layer, the prediction artifacts for both benchmarks, and the generated
reports accompany this submission; Appendix~\ref{app:repro} gives the exact
commands that regenerate every number and figure in this paper.

\section{Results}
\label{sec:results}

\subsection{A curated benchmark still carries an excess zero-gain rate}
\label{sec:res-textonly}

\begin{figure}[t]
  \centering
  \includegraphics[width=\linewidth]{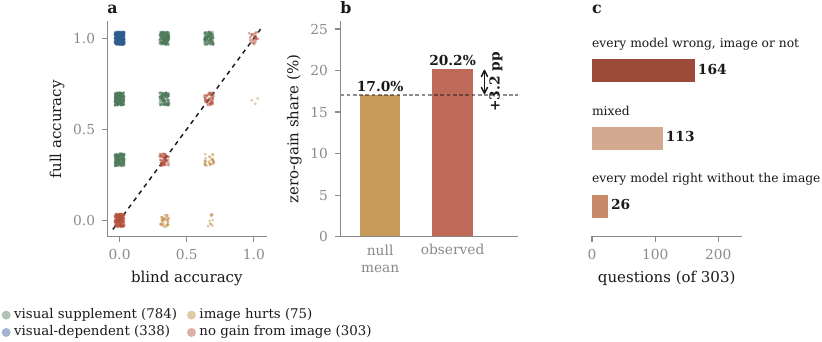}
  \caption{The zero-gain signal on MMStar. \textbf{(a)}~Model-averaged full
  against blind accuracy per question (judged arm, jittered; three models
  give the $4\times4$ grid). Questions on the diagonal gain nothing from the
  image. \textbf{(b)}~The observed 20.2\% share against 999 item-shuffled
  permutations: the null centres at 17.0\%, leaving a $+3.2$\,pp excess that
  no permutation reaches ($p=0.001$). \textbf{(c)}~The no-gain set is
  heterogeneous: on 164 of 303 questions every model is wrong with and
  without the image, so the conservative reading is \emph{zero aggregate
  accuracy gain from the image for this model pool}, and genuinely
  text-answerable items are only one possible component.}
  \label{fig:vd}
\end{figure}

Figure~\ref{fig:vd} and Table~\ref{tab:pools} give the headline result. The
observed zero-gain share is 20.2\% (Wilson 95\% CI $[18.2, 22.3]$); the
permutation null predicts 17.0\%; the excess is $+3.2$\,pp at $p=0.001$, the
floor of a 999-permutation test. The excess survives every
leave-one-model-out pool ($+3.6$ to $+7.7$\,pp, each $p=0.001$).
Across 20 random split-half checks, the median absolute difference between
the two half-sample excess estimates is 1.5\,pp.

Two readings of this result must be kept apart (Figure~\ref{fig:vd}c). Of
the 303 zero-gain questions, 164 are items on which every model is wrong
both with and without the image, which is evidence of shared failure.
Only 26 are answered correctly by all three models blind.
Our claim is therefore limited to equal model-averaged full and blind
accuracy for this pool. The excess quantifies how often that equality occurs
beyond marginal-rate arithmetic; it does not identify which underlying
mechanism produced it. The modest $+3.2$\,pp excess is also consistent with
MMStar's curation having removed most non-visual shortcuts.

\begin{table}[t]
  \caption{Zero-gain share against the permutation null on MMStar (999
  permutations, one-sided, $(g{+}1)/(N{+}1)$ correction). The excess is
  markedly more pool-stable than the raw share: raw spread 14.2\,pp, excess
  spread 4.6\,pp.}
  \label{tab:pools}
  \centering
  \small
  \begin{tabular}{lcccc}
    \toprule
    Model pool & Observed (\%) & Null mean (\%) & Excess (pp) & $p$ \\
    \midrule
    All three models        & 20.2 & 17.0 & $+3.2$ & 0.001 \\
    without InternVL3-8B    & 27.9 & 24.1 & $+3.7$ & 0.001 \\
    without Ristretto-3B    & 25.5 & 21.9 & $+3.6$ & 0.001 \\
    without gpt-5-nano      & 34.4 & 26.7 & $+7.7$ & 0.001 \\
    \bottomrule
  \end{tabular}
\end{table}

\subsection{The two audit signals have different category profiles}
\label{sec:res-strata}

\begin{figure}[t]
  \centering
  \includegraphics[width=\linewidth]{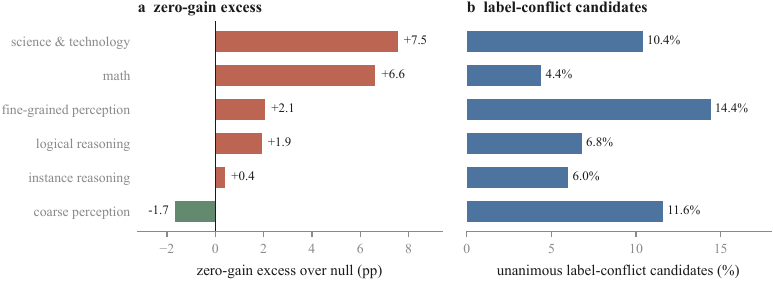}
  \caption{Per-category audit of MMStar (250 items per category), sorted by
zero-gain excess. \textbf{(a)}~The excess concentrates in mathematics and
science \& technology and is absent from coarse perception, a pattern
consistent with greater scope for non-visual knowledge in the former
categories. \textbf{(b)}~Unanimous label-conflict candidates peak in the
perception categories. The two detectors localize different candidate
failure modes.}
  \label{fig:strata}
\end{figure}

Stratifying by MMStar's six official categories turns one number into a
pattern (Figure~\ref{fig:strata}). The zero-gain excess is largest in math
($+6.6$\,pp) and science \& technology ($+7.5$\,pp), where non-visual
knowledge is a plausible explanation. Coarse perception shows no positive
excess ($-1.7$\,pp). Unanimous label-conflict candidates distribute
differently: they peak in the perception categories (11.6\% coarse,
14.4\% fine-grained) and are lowest in math and reasoning (4.4--6.8\%), with
science \& technology intermediate (10.4\%). These category profiles show
that the two detectors capture distinct signals; neither profile by itself
identifies the causal mechanism behind its signal.

\paragraph{Blind-run abstentions.}
The extraction discrepancy is distinct from refusal under the blind
ablation. Full/blind unparseable rates are 0.0/20.5\% for InternVL3-8B,
0.0/1.1\% for Ristretto-3B, and 10.3/67.5\% for gpt-5-nano. Consequently,
zero-gain and visual-gain categories reflect both benchmark properties and
model-specific willingness to answer without an image. We retain all three
models in the primary pool and expose this dependence through the
leave-one-model-out analysis; results should not be extrapolated to pools
with different refusal policies.

\subsection{Label-conflict candidates}
\label{sec:res-consensus}

The consensus detector flags 23.9\% of MMStar items at some tier; we focus
on the unanimous tier. On 134 questions (8.9\%, Wilson 95\% CI
$[7.6, 10.5]$) every parseable model answer is the same option and it is not
the gold label; on 117 of those all three models answered in parseable form,
and on the remaining 17, only two models produced parseable answers because
abstentions are excluded from the vote. On the 1{,}346 items for which all
three models produced parseable option
answers, complete-case Fleiss' $\kappa$ is 0.544 (item-bootstrap 95\% CI
$[0.517, 0.570]$), indicating moderate agreement beyond that expected from
the models' marginal answer distributions. Because the excluded 154 items
are non-random unparseable outputs from gpt-5-nano, this estimate describes
the complete-case subset rather than all 1{,}500 MMStar items. Unanimous
agreement therefore stands well apart from the pool's baseline
behavior. The rate is of the same order as the
annotation-error rates reported for MMStar by the independent fallibility
analysis discussed in Section~\ref{sec:related}, which did collect fresh
human labels.

We deliberately call these \emph{candidates}, and the reason is measurable
in our own pool. Errors of different models are correlated
(Section~\ref{sec:related}): when two of our models are both wrong, they pick
the same wrong option in 64--74\% of cases, against 26--27\% when the
item-to-item pairing is destroyed by the same shuffle that defines our null.
Unanimity therefore serves as a prioritization signal: it
concentrates human review, which remains the arbiter. The audit layer
releases the flagged list with per-question model votes for exactly that
purpose.

\subsection{Removing the LLM judge changes almost nothing, for a reason}
\label{sec:res-judge}

\begin{figure}[t]
  \centering
  \includegraphics[width=\linewidth]{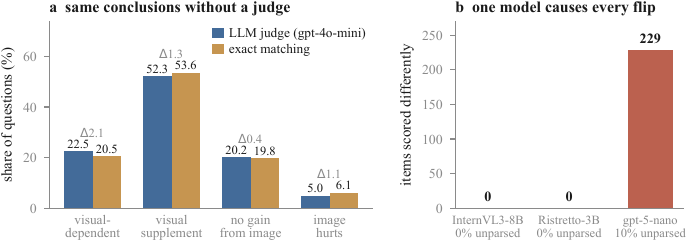}
  \caption{Judge dependence of the audit on MMStar. \textbf{(a)}~Category
  shares under the \texttt{gpt-4o-mini} judge and under exact letter
  matching, on identical predictions: every share moves by at most 2.1\,pp
  and no conclusion changes. \textbf{(b)}~The disagreement has a single
  source: for the two open models the judge and the letter parser score
  every item identically, while gpt-5-nano, whose answers are unparseable on
  10\% of items, accounts for all of it.}
  \label{fig:robust}
\end{figure}

Audit claims should not silently inherit an LLM judge. Four of six
detectors never use one
(Table~\ref{tab:detectors}). For the two that do, we recomputed everything
under exact letter matching on identical predictions
(Figure~\ref{fig:robust}a): category shares move by at most 2.1\,pp
(zero-gain 19.8\% against 20.2\%; visual-dependent 20.5\% against 22.5\%;
supplement 53.6\% against 52.3\%; conflicting 6.1\% against 5.0\%), and
nothing in Sections~\ref{sec:res-textonly}--\ref{sec:res-consensus} changes.

The instructive part is where the two arms disagree. 122 of 1{,}500 items
(8.1\%) change visual-dependency category between arms, and every one of
them traces to gpt-5-nano: the judge and the letter parser score that model
differently on 229 model-item pairs, and the two open models on none at all
(Figure~\ref{fig:robust}b). The mechanism is prosaic: gpt-5-nano answers in
free-form prose on 10\% of items, which a regular expression cannot read but
a judge can. The judge is thus a robustness arm: for models with canonical
outputs exact matching suffices, and for models without canonical outputs
the audit report can say so per model instead of averaging the discrepancy
away.

\subsection{Where the detectors stop working: a binary-format probe}
\label{sec:res-binary}

We ran the same audit on HallusionBench, whose
951 items are yes/no questions, to measure how the detectors behave outside
their design format. With
two answer options blind guessing sits at 50\%, unanimity is easy to obtain
by chance, and the
multiple-choice consensus machinery does not apply at all: option letters do
not parse and the exact arm reports abstention everywhere. The zero-gain
excess is still large and significant ($+5.4$\,pp over an 18.5\% null,
$p=0.001$), but the decisive calibration test fails. Ranking items by their
continuous visual gain recovers HallusionBench's own visual-dependent
against visual-supplement design labels at AUC 0.49, that is, at chance
(Appendix~\ref{app:binary}). One directional signal survives and matches the
benchmark's intent: the excess is approximately 2.2 times larger on design-VS items
($+7.6$\,pp) than on design-VD items ($+3.4$\,pp). We conclude that the
current evidence supports use on the four-option MMStar setting evaluated
here and places binary formats outside the demonstrated scope. Calibration
across other $k$-way multiple-choice benchmarks remains to be measured.

\subsection{Auditing the audit}
\label{sec:res-selfaudit}

Audit pipelines can manufacture their own findings, so we checked ours
against the known failure modes. \emph{No-op perturbation}: the blind run
strips images centrally, blind transcripts contain model statements that no
image is present, and blind accuracy falls towards the option prior in the
perception categories. \emph{Extraction artifacts}: the judged-against-exact
arm (Section~\ref{sec:res-judge}) measures precisely this and attributes all
discrepancy to one model. \emph{Unfaithful scoring}: the judged arm is
VLMEvalKit's native protocol, unmodified. \emph{Broken pairing in
uncertainty}: the permutation null is item-paired by construction, and only
the full-to-blind linkage is destroyed. \emph{One scalar, two readings}:
every headline is disaggregated (category strata, per-pool tables, and the
164-of-303 composition in Figure~\ref{fig:vd}c). One bug we did catch and
fix is reported in Section~\ref{sec:method-inputs}: back-filling gold labels
into model answers had suppressed unanimous conflicts from 134 to 12, and
the layer now refuses to do it.

\section{Limitations}
\label{sec:limitations}

\emph{Model pool.} A pool of three models fixes the audit's signals but
leaves the exact category shares pool-dependent; we treat it as a working
approximation whose sensitivity is measured rather than assumed. The direction and
significance of the headline excess are unchanged in every
leave-one-model-out pool (Table~\ref{tab:pools}), category shares are
published with their pool ranges, and each model's blind-refusal profile
is reported alongside (Section~\ref{sec:res-strata}). A larger pool will
tighten the per-item accuracy grid and is part of ongoing work.
\emph{Mechanism attribution.} The zero-gain excess is an aggregate,
mechanism-agnostic quantity: it bounds the combined effect of language
priors, option statistics and contamination, and the category strata
localize it without singling out one cause.
\emph{Scope.} We audit one benchmark in depth under a four-option
multiple-choice format. The binary-format boundary is measured in
Section~\ref{sec:res-binary}, and base rates are benchmark-specific by
construction, so applying the audit to a new suite means rerunning it
there.

\section{Conclusion}
\label{sec:conclusion}

In this work we ask how far the artifacts of a completed evaluation can be
pushed as an audit of the benchmark that produced them. We add an audit
layer to VLMEvalKit: six detectors that run as post-processing over the
saved per-question outputs, one text-only rerun, and a statistics module
that reports each finding as an excess over an item-matched permutation
null. Auditing MMStar with three models, we measure a zero-gain rate of
20.2\%, of which $+3.2$\,pp exceeds the null ($p=0.001$) and concentrates
in the mathematics and science categories, and we flag 134 questions whose
parseable answers agree on a non-gold option, concentrated instead in the
perception categories. We also measure where the audit stops working: on a
binary yes/no benchmark the detectors do not recover the benchmark's own
design labels. Practitioners read a gap between leaderboard scores as a
difference in capability; our findings advise caution when that gap is
narrower than the artifact rates measured on the same benchmark, and they
suggest reporting any claim about a benchmark's defects together with its
null, its interval and its model-pool range. Three models fix the
direction of both signals, and their magnitudes remain pool-dependent; the
134 candidates stay candidates until human review turns them into a
measured error rate. We release the audit layer, the prediction artifacts
and the per-question reports, and invite the community to audit the
benchmarks it relies on.

\bibliographystyle{plainnat}
{\small
\bibliography{references}}

\appendix

\section{Reproducing every number in this paper}
\label{app:repro}

The audit consumes only the per-model result files that VLMEvalKit writes
for a normal evaluation, plus the matching blind-run files. For each benchmark,
the following two commands regenerate its statistics
and figures; the HallusionBench invocation additionally passes
\texttt{--design-prefix VD} to both scripts.

\begin{quote}\small\ttfamily
python scripts/audit\_significance.py \textbackslash\\
\hspace*{1em}--full  <model>\_MMStar\_gpt-4o-mini\_result.xlsx (one per model) \textbackslash\\
\hspace*{1em}--blind <model>\_MMStar\_gpt-4o-mini\_result\_blind.xlsx \textbackslash\\
\hspace*{1em}--models InternVL3-8B Ristretto-3B gpt-5-nano \textbackslash\\
\hspace*{1em}--strata-column category --perms 999 --seed 7 --out <report dir>
\end{quote}

\begin{quote}\small\ttfamily
python scripts/audit\_paper\_figures.py \textbackslash\\
\hspace*{1em}--full ... --blind ... --models ... --stats-json ... --out paper/figures
\end{quote}

The blind artifacts themselves come from a normal evaluation invocation with
the \texttt{--blind} flag, which is the only step that requires inference:

\begin{quote}\small\ttfamily
python run.py --data MMStar --model <model> --blind
\end{quote}

The statistics script emits the canonical JSON report. The figure script
reads all reported aggregate and stochastic values from that JSON and uses
the evaluation files only for deterministic per-item panels; it does not
rerun the permutation null. The scripts depend only on
\texttt{pandas}/\texttt{openpyxl} (plus \texttt{matplotlib} for figures)
and do not import the toolkit itself. Randomness enters only when the
canonical report is generated, through the recorded permutation seed.

\paragraph{Answer extraction (exact arm).} A prediction is mapped to an
option letter by three ordered rules: a bare letter, possibly punctuated; a
letter followed by a delimiter at the start of the string; or a letter
following the words ``answer'' or ``option''. Anything else becomes the
abstention sentinel \texttt{Z}, which is excluded from consensus votes.
This mirrors the toolkit's own extractor, so the audit does not introduce a
second, divergent notion of what a model answered.

\paragraph{Permutation null.} For each of 999 permutations, one random
permutation of question IDs is drawn and applied to the blind hit vectors of
\emph{all} models simultaneously. This preserves each model's marginal blind
accuracy and the correlation between models' blind answers, and destroys
only the per-question pairing between full and blind runs. The zero-gain
share is recomputed on each permuted dataset with the identical category
rule; the reported excess is observed minus null mean, and the one-sided
$p$-value is $(g{+}1)/(N{+}1)$, where $g$ counts permutations with a share
at least as large as observed. Stratified nulls repeat the procedure within
each category. For the split-half check, questions are randomly halved and the excess is
computed in each half. Across 20 resplits, the median absolute difference
between half-sample estimates is 1.5\,pp on MMStar and 2.1\,pp on
HallusionBench; these values are stability diagnostics.

\section{Threshold sensitivity of the visual-dependent share}
\label{app:sweep}

The visual-dependent share depends strongly on its definition, which is why
the paper never headlines it. On MMStar (judged arm), the strict all-or-none
definition flags 338 items (22.5\%), while a majority definition (at least
two of three models correct with the image, at most one without) flags 822
items (54.8\%). On HallusionBench the same pair is 140 (14.7\%) against
478 (50.3\%). The text-only category has no threshold at $n=3$, since the
accuracy grid makes it exact equality, and the unanimous consensus count has
none at any $n$; those are the quantities the paper reports.

\section{Binary-format calibration check}
\label{app:binary}

\begin{figure}[h]
  \centering
  \includegraphics[width=\linewidth]{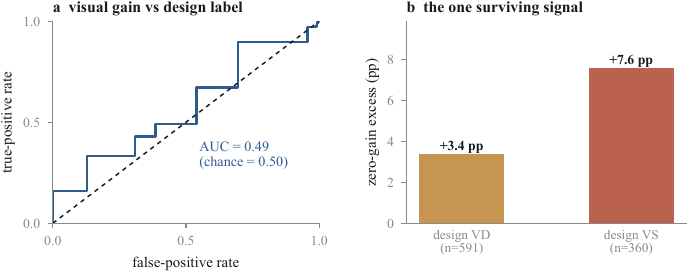}
  \caption{HallusionBench (951 yes/no items, same three models).
  \textbf{(a)}~Ranking items by the continuous visual gain does not recover
  the benchmark's own design labels (AUC 0.49 against 0.50 for chance).
  \textbf{(b)}~The one directional signal that survives is consistent with
the benchmark's design: zero-gain excess is approximately 2.2 times larger
on visual-supplement items than on visual-dependent items.}
  \label{fig:binary}
\end{figure}

Design labels are read from the question index prefix that HallusionBench
itself assigns (\texttt{VD} for visual-dependent items, the remainder
treated as visual-supplement), giving 591 positives and 360 negatives. The
AUC is computed on the continuous per-question visual gain
$a^F_q - a^B_q$, with ties counted as half wins. Together with the
degenerate letter-extraction behaviour described in
Section~\ref{sec:res-binary}, this is the evidence for excluding binary
formats from the detectors' current calibration.

\end{document}